\documentclass[10pt,twocolumn,letterpaper]{article}
\usepackage{iccv}
\usepackage{times}
\usepackage{graphicx}
\usepackage{amsmath}
\usepackage{amssymb}
\usepackage{amsthm}
\usepackage{apxproof}
\usepackage{booktabs}
\usepackage{caption}
\usepackage{capt-of}
\usepackage{cuted}
\usepackage{subcaption}
\usepackage[font=small,labelfont=bf,tableposition=top]{caption}
\usepackage[pagebackref=true,breaklinks=true,letterpaper=true,colorlinks,bookmarks=false]{hyperref}
\usepackage{cleveref}
\usepackage[accsupp]{axessibility}  

\newcommand{\x}{\boldsymbol{x}}
\newcommand{\y}{\boldsymbol{y}}

\newcommand{\bv}{\boldsymbol{v}}
\newcommand{\bc}{\boldsymbol{c}}
\newcommand{\beps}{\boldsymbol{\varepsilon}}

\makeatletter
\renewcommand{\paragraph}{%
\@startsection{paragraph}{4}%
{\z@}{0.25em}{-1em}%
{\normalfont\normalsize\bfseries}}
\makeatother

\iccvfinalcopy
\def\iccvPaperID{10150}
\ificcvfinal\fi

\title{Viewset Diffusion: (0-)Image-Conditioned 3D Generative Models from 2D Data}

\author{Stanislaw Szymanowicz \quad Christian Rupprecht \quad Andrea Vedaldi\\[0.3em]
Visual Geometry Group --- University of Oxford\\
{\tt\small \{stan,chrisr,vedaldi\}@robots.ox.ac.uk}}

\begin{document}

\maketitle

\ificcvfinal\thispagestyle{empty}\fi

\begin{strip}\centering
\includegraphics[width=\textwidth]{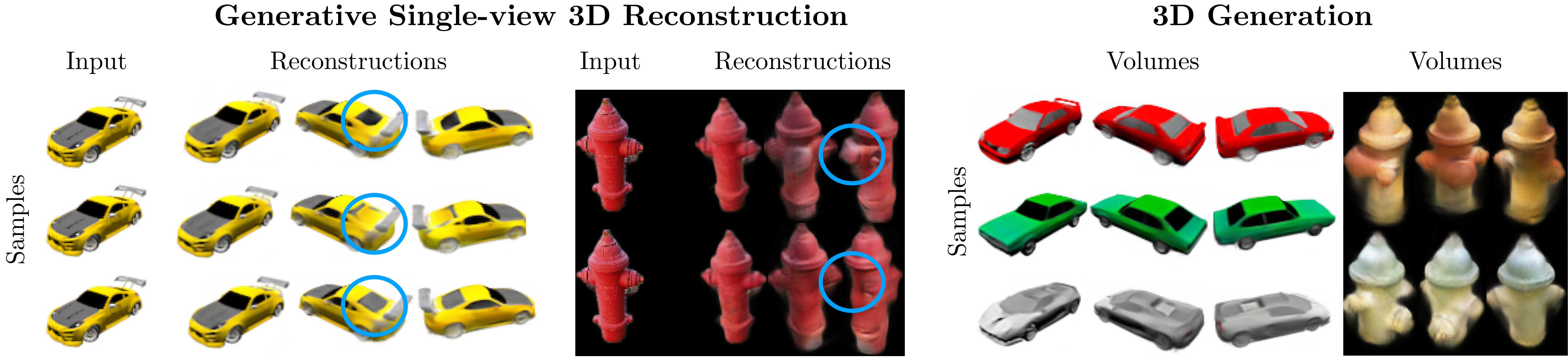}
\captionof{figure}{
\textbf{Viewset Diffusion.} Our category-specific models perform both `generative' 3D reconstruction and unconditional 3D generation.
In single-view 3D reconstruction (left) our models generate plausible explanations of occluded regions (car's back, hydrant's occluded side).
The same models are able to generate varied 3D objects (right) in a feed-forward manner while being trained only on 2D data.%
\label{fig:teaser}}
\end{strip}

\begin{abstract}
We present \emph{Viewset Diffusion}, a diffusion-based generator that outputs 3D objects while only using multi-view 2D data for supervision.
We note that there exists a one-to-one mapping between viewsets, \ie, collections of several 2D views of an object, and 3D models.
Hence, we train a diffusion model to generate viewsets, but design the neural network generator to reconstruct internally corresponding 3D models, thus generating those too.
We fit a diffusion model to a large number of viewsets for a given category of objects.
The resulting generator can be conditioned on zero, one or more input views.
Conditioned on a single view, it performs 3D reconstruction accounting for the ambiguity of the task and allowing to sample multiple solutions compatible with the input.
The model performs reconstruction efficiently, in a feed-forward manner, and is trained using only rendering losses using as few as three views per viewset.
Project page: \url{szymanowiczs.github.io/viewset-diffusion}.
\end{abstract}

\section{Introduction}%
\label{s:introduction}

Image-based 3D reconstruction, \ie, recovering the 3D shape of the world from 2D observations, is a fundamental problem in computer vision.
In this work, we study the problem of reconstructing the 3D shape and appearance of individual objects from as few as one image.
In fact, we cast this as image-conditioned 3D generation, and also consider the case of unconditional generation (\cref{fig:teaser}).

Single-view 3D reconstruction is inherently ambiguous because projecting a 3D scene to an image loses the depth dimension.
The goal is thus not to recover the exact 3D shape and appearance of the object, particularly of its occluded parts, but to generate \emph{plausible} reconstructions.
This can only be achieved by learning a prior over the likely 3D shapes and appearances of the objects.
Here, we do so for one category of objects at a time.

\begin{figure}
\centering
\includegraphics[width=\columnwidth]{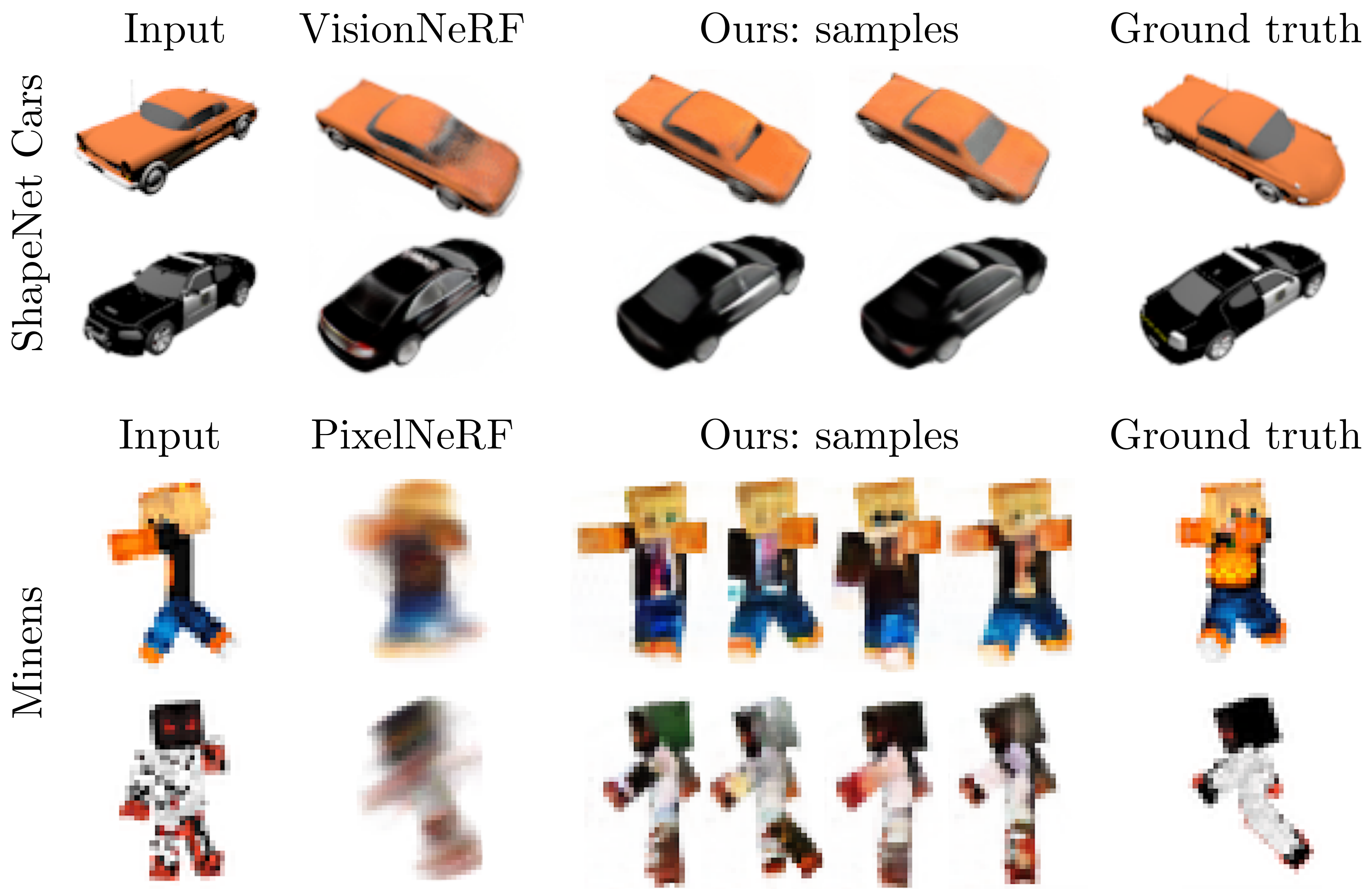}
\caption{\textbf{Ambiguities.} Under occlusion, deterministic methods blur possible shapes (orange car's back, Minens characters' poses) and colours (black car's back, occluded sides of Minens characters). Our method samples plausible 3D reconstructions.}%
\label{fig:ambiguities}
\end{figure}

Leveraging 3D object priors for reconstruction has been explored by several works~\cite{gwak2017weakly,wu2018shapepriors}.
Most of these tackle 3D reconstruction in a deterministic manner, outputting one reconstruction per object.
This is limiting in the presence of ambiguity, as a deterministic reconstructor can only predict either (1) a single most likely solution, which is plausible but usually incorrect, or (2) an average of all possible reconstructions, which is implausible (\cref{fig:ambiguities}).

Thus, in this work, we tackle the problem of modelling ambiguity in few-view 3D reconstruction.
Our goal is to learn a conditional generator that can sample \emph{all plausible 3D reconstructions} consistent with a given image of an object from a given viewpoint.

We approach this problem using Denoising Diffusion Probabilistic Models (DDPM)~\cite{ho20denoising} due to their excellent performance for image generation~\cite{dhariwal21diffusion}.
However, while DDPMs are trained on billions of images, 3D training data is substantially more scarce.
We thus seek to \emph{learn a 3D DDPM using only multi-view 2D data} for supervision.
The challenge is that DDPMs assume that the training data is in the same modality as the generated data.
In our setting, the 3D model of the object can be thought of as an unobserved \emph{latent variable}, learning which is beyond the scope of standard DDPMs.
We solve this problem starting from the following important observation.

Given a 3D model of an object, we can render all possible 2D views of it.
Likewise, given a sufficiently large set of views of the object, called a \emph{viewset}, we can recover, up to an equivalence class, the corresponding 3D model.
Because of this bijective mapping, generating 3D models is equivalent to generating viewsets.
The advantage of the latter is that we often have access to a source of suitable 2D multi-view data for supervision.
Similarly to 2D image generation, our corresponding DDPM takes as input a partially noised viewset and produces as output a denoised version of it.
For generation, this denoising process is iterated starting from a Gaussian noise sample.

\begin{figure}
\centering
\includegraphics[width=\columnwidth]{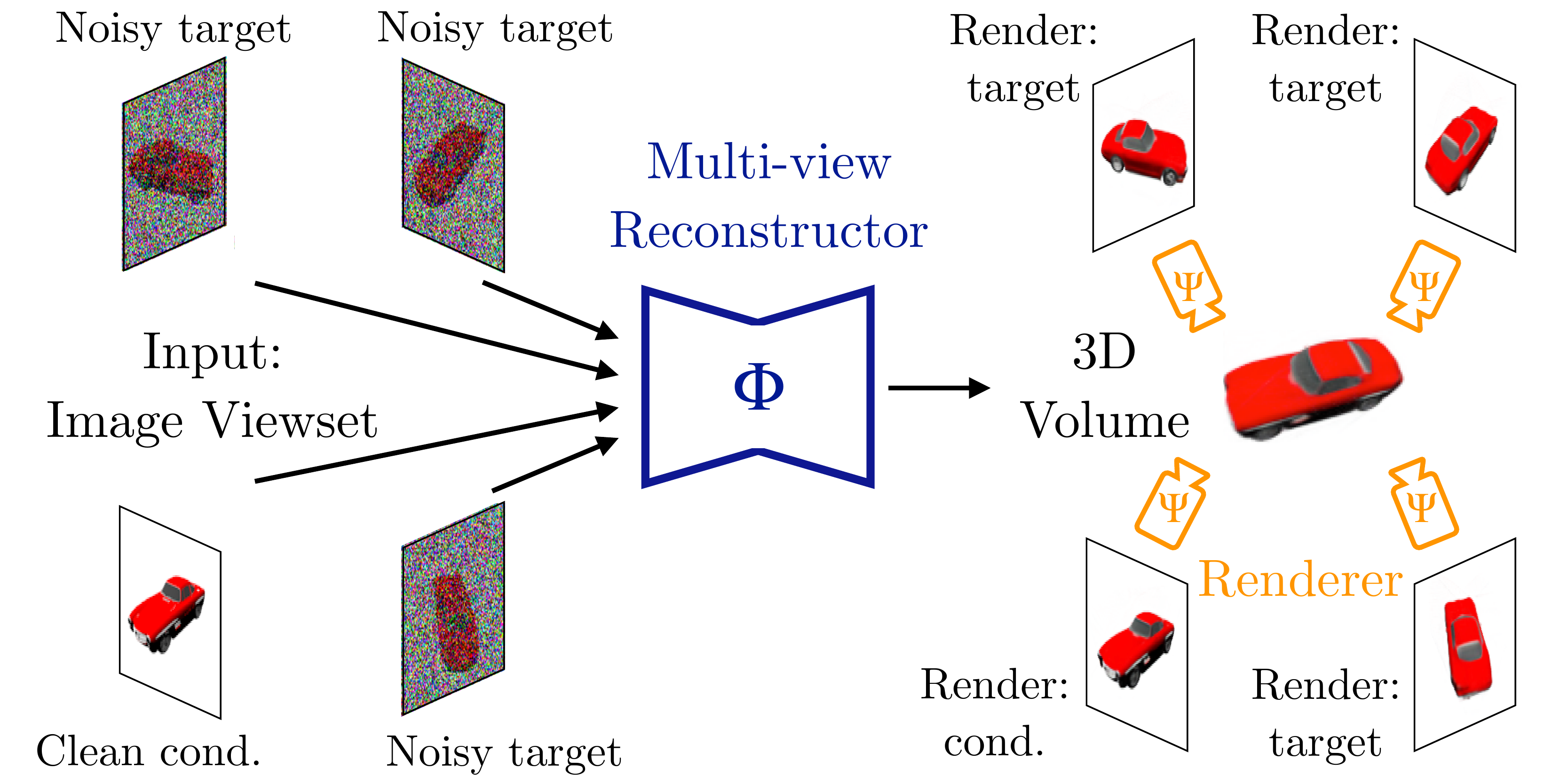}
\caption{\textbf{Viewset Diffusion} takes in \emph{any} number of clean conditioning images and generated images with Gaussian noise (\cref{s:method_viewset}).
The denoising function is defined as reconstructing (\cref{s:method_network}) and rendering a 3D volume.
When there is at least one clean conditioning view, Viewset Diffusion samples plausible 3D reconstructions.
When all input views are noisy, it samples a 3D object form the prior.}%
\label{fig:method}
\end{figure}

Our second key intuition is that the bijective mapping between viewsets and 3D models \emph{can be integrated in the denoising network} itself.
Namely, our DDPM is designed to denoise the input viewset by reconstructing a full radiance field of the corresponding 3D object (see \cref{fig:method}).
This has the advantage of producing the 3D model we are after and ensuring that the denoised viewset is 3D consistent (the lack of 3D consistency is an issue for some multi-view generators~\cite{liu2023zero1to3,watson20223dim}).
Furthermore, by allowing different views in the viewset to be affected by different amounts of noise, the same model supports conditional generation from any number of input views (including zero).
This conditional generation is achieved by setting the noise level of the available conditioning images to zero.

We call our method \emph{Viewset Diffusion} and, with it, make several contributions:
(i) The idea of generating viewsets as a way to apply DDPMs to the generation of 3D objects even when only multi-view 2D supervision is available.
(ii) An ambiguity-aware 3D reconstruction model that is able to sample different plausible reconstructions given a single input image, and which doubles as an unconditional 3D generator.
(iii) A network architecture that enables our reconstructions to match the conditioning images, aggregate information from an arbitrary number of views in an occlusion-aware manner and estimate plausible 3D geometries.
(iv) A new synthetic benchmark dataset, designed for evaluating the performance of single-image reconstruction techniques in ambiguous settings.

\section{Related Work}\label{s:related}

Works most related to ours consider the problem of few-view 3D reconstruction and the ensuing ambiguities.

\paragraph{Reconstructing Neural Fields.}

Few-view reconstruction methods that use Neural Radiance Fields~\cite{mildenhall20nerf:} include Sin-NeRF~\cite{xu22sinnerf:} and DietNeRF~\cite{jain2021dietnerf}.
They use semantic pseudo-labels in unseen views to provide multi-view pseudo-constraints from pre-trained 2D networks, effectively leveraging a 2D prior.
Other works learn a 3D prior instead and represent individual shapes using global latent codes~\cite{jang21codenerf:,muller22autorf:,rematas21sharf:,sitzmann2021lfns}.
Codes be optimised at test-time via auto-decoding, akin to latent space inversion in priors learned with 3D GANs~\cite{chan2021eg3d}.
The latent codes can also be local~\cite{henzler21unsupervised, lin2023visionnerf, yu21pixelnerf:} or simultaneously global and local~\cite{lin2023visionnerf}, which tends to improve high-frequency details.
While our method borrows the idea of local conditioning and learning a 3D prior from such prior works, a key difference is that we \emph{sample different plausible reconstructions}, while most prior works only output a \emph{single reconstruction}, which tends to regress to the mean, usually falling outside the data distribution, being blurry, and mixing several modes in one.
In contrast, our method samples multiple sharp reconstructions, each of which is different yet plausible.

\paragraph{Reconstruction Beyond Neural Fields.}

Many other possible 3D representations have been explored, including geometry-free representations~\cite{reizenstein21co3d, suhail2022generalizable, suhail2022lfnr, wang2021ibrnet}, occupancy voxel grids grids~\cite{choy20163d,  Tulsiani_2017_CVPR, xinchen2016perspective_transfomer}, textured meshes~\cite{kanazawa2018cmr, wu2021dove} or hybrid implicit-explicit representations~\cite{shen2021dmtet,wu2023magicpony}.
While our work is currently based on a neural radiance field, it is compatible with any differentiable formulation.

\paragraph{Ambiguity in 3D Reconstruction.}

Single-view 3D reconstruction is an ill-posed problem because a 2D input only partially constrains the 3D output.
A 3D prior can reduce but not eliminate the ambiguity.
In fact, even when the 3D reconstruction is constrained to be plausible according to the given prior, it does not mean that it is unique~\cite{wu2018shapepriors}.
For instance, when we reconstruct a person seen from the front, even knowing that it is a person is insufficient to exactly predict their back.
The goal is to obtain one or several plausible solutions that are compatible with the given observations, for example via constrained optimization~\cite{gwak2017weakly} or using an adversarial loss during training~\cite{wu2018shapepriors}.
Ignoring ambiguity may result in distortions, including blurry reconstructions without fine details~\cite{dai2017complete}.

We embrace ambiguity in 3D reconstruction and train a network to
(1) output a 3D object that matches an observation such that
(2) the 3D object is a plausible member of a given category.
This setting is similar to Wu~\etal~\cite{wu2018shapepriors}, but we allow sampling different plausible reconstructions via a conditional diffusion model, rather than finding a single plausible solution via a constrained optimisation approach.

\paragraph{3D Modelling with Diffusion Models}

Recently, Denoising Diffusion Probabilistic Models~\cite{ho20denoising} (DDPM) have been applied to modelling 3D shape distributions by diffusing directly in the space of 3D representations, including point clouds~\cite{luo2021diffusion}, triplanes~\cite{gupta233dgen:,shue2022nfd,wang2023rodin} and other radiance fields~\cite{jun23shape-e:,mueller2022diffrf}
Shortcomings of these approaches include (1) assuming an available 3D dataset (see also \cref{s:method_challenge}) and (2) requiring heuristics for dealing with `floater' artefacts~\cite{mueller2022diffrf,shue2022nfd} common in volumes reconstructed from multi-view data in an optimisation setting.

Others leverage pre-trained 2D diffusion models via Score Distillation Loss for text-to-3D generation~\cite{poole2022dreamfusion,wang22jacobianchaining} with test-time optimisation, and extensions allow image-based reconstruction~\cite{li20223ddesigner,melas2023realfusion}.
Concurrently to our work, several others~\cite{chan2023genvs,liu2023zero1to3,watson20223dim} learn image-conditioned diffusion models.
The outputs are 2D, and their 3D consistency is only approximate, with frequent flickers~\cite{chan2023genvs,liu2023zero1to3,watson20223dim}.
3D consistency can be enforced via costly (sometimes 1 hour~\cite{zhou2022sparsefusion}) test-time optimization of a 3D representation~\cite{gu2023nerfdiff,zhou2022sparsefusion}.
In contrast to prior and concurrent works, our method
(1) can be trained on \textbf{2D data}, requiring only 3 views of an object,
(2) is guaranteed to be 3D consistent, and 
(3) is feed-forward, and therefore much faster than test-time distillation methods~\cite{gu2023nerfdiff,melas2023realfusion,zhou2022sparsefusion}.
HoloDiffusion~\cite{karnewar2023holodiffusion} also learns 3D generative models from 2D data, but it only considers unconditional generation, while we propose a principled, unified framework for conditional and unconditional generation.

The closest work to ours is RenderDiffusion~\cite{anciukevicius2022renderdiffusion}, discussed in detail in \cref{s:method_discussion}.

\section{Method}\label{s:method}

We consider the problem of learning a distribution over 3D objects, supporting both unconditional sampling and sampling conditioned on one or more views of the object.
We approach this problem using a 3D DDPM and rethink the training setup to allow training it from multi-view 2D data, without access to 3D ground-truth.

\subsection{DDPMs: background and notation}

Consider the problem of learning a distribution $p(x)$ over 2D images $x \in \mathbb{R}^{3\times H \times W}$ (or, with little changes, a distribution $p(x|y)$ conditioned on additional information $y$ such as a text description).
The DDPM approach generates a sequence of increasingly noisier versions of the data.
This sequence starts from $x_0 = x$ and adds progressively more Gaussian noise such that the conditional distribution $p(x_t | x_0)$ at step $t$ can be characterised by writing $x_t = \sqrt{1 - \sigma_t^2} x_0 + \sigma_t \epsilon_t$, $t=1,\dots,T$, where $\sigma_t$ is a sequence of noise standard deviations increasing from $0$ to $1$, $\epsilon_t$ is normally distributed.
The marginal distribution $p(x_t)$ does not, in general, have a closed-form solution but for large $t \approx T$ it approaches a normal distribution.

In order to draw a sample $x_0$, one starts backwards, drawing first a sample $x_T$ from the marginal $p(x_T)$, and then taking samples $x_t$ from $p(x_{t-1}|x_t)$, until $x_0$ is obtained.
The key observation is that these are comparatively simple distributions to learn.
Various slightly different formulations are possible;
here, we learn a denoising network $\hat x_0(x_t,t)$ that tries to estimate the ``clean'' sample $x_0$ from its noisy version $x_t$.
Given a training set $\mathcal{X}$ of images, such a network is trained by minimizing the loss
$$
\mathcal{L}(\hat x_0,t)
=
\frac{1}{|\mathcal{X}|}
\sum_{x_0\in \mathcal{X}} w(\sigma_t) \,
\mathbb{E}_{p(x_t|x_0)}
\| \hat x_0(x_t,t) - x_0 \|^2
$$
where the weight $w(\sigma_t)$ depends on the noise/timestep~\cite{hang2023minsnr}.

\subsection{The challenge of a 3D extension}\label{s:method_challenge}

We now consider using a DDPM to learn a distribution $p(\bv)$ of 3D models $\bv$ of objects (in practice radiance fields, see \cref{s:method_network}).
In oder to train a DDPM to generate 3D models $\bv$, we would require a dataset $\mathcal{V}$ of such models.
Differently from 2D images, however, 3D models are not readily available.
We assume instead to have access to 2D multi-view training data.
Each training sample is a \emph{viewset} $(\x, \Pi)$, \ie, a collection $\x \in \mathbb{R}^{N\times 3\times H\times W}$ of $N$ views of a 3D object with known camera poses $\Pi = \{\pi^{(i)}\}_{i=1}^N$.
The 3D model $\bv$ is not observed but a \emph{latent variable}.

The fact that $\bv$ is a latent variable suggests adopting the \emph{latent diffusion} approach, which has been very successful for 2D images~\cite{rombach22high-resolution}, and thus simply replace the input data $\x$ with corresponding codes $\bv$.
Unfortunately, doing so requires to know the encoder $\x\mapsto\bv$, mapping the input data $\x$ to the latents $\bv$.
In our case, this mapping amounts to image-based 3D reconstruction, which is non-trivial.

One way of implementing the mapping $\x\mapsto\bv$ is to use an optimization method like NeRF, which can recover the radiance field $\bv$ given a sufficiently large viewset $\x$.
This is the approach taken by several prior works~\cite{gupta233dgen:,wang2023rodin}, but it has several shortcomings.
First, with no prior information, reconstructing a model $\bv$ from a viewset $\x$ is ambiguous due to visibility (the interior of an object does not matter to its appearance) and over-parameterization.
While we can dismiss these ambiguities as classes of equivalent models\footnote{In the sense that these models all produce the same images.}, nevertheless they increase the irregularity of the distribution $p(\bv)$ that we wish to learn~\cite{wang2023rodin}.
Second, good independent reconstructions require  a fairly large ($\ge 50$) number of views per object, which are not always available, and even then they may still contain defects such as floaters~\cite{shue2022nfd}.
Lastly, optimization-based reconstruction is slow (hours per sample) and must happen for every sample $\x\in\mathcal{X}$ before training the distribution $p(\bv)$ can even start.

\subsection{Viewset diffusion}\label{s:method_viewset}

In this section, we seek to directly train a DDPM to generate 3D models using only 2D supervision.
Our approach is centred around a few simple but powerful observations.
First, we note that it is easy to apply DDPMs to viewsets $\x$ because, differently from the 3D radiance fields $\bv$, they are assumed to be observable.
Second, while DDPMs do not directly support generating latent variables, we can interpret the latent 3D model $\bv$ as an \emph{intermediate viewset representation} learned by the {neural network} which implements the DDPM --- we simply do not apply diffusion to it.

Concretely, we write the DDPM denoising function as the composition of two functions.
The first is an encoder network
\begin{equation}\label{e:encoder}
\bv = \Phi(\x_t, \Pi, \sigma_t)
\end{equation}
which, given a \emph{noised viewset} $(\x_t,\Pi)$, produces as output a 3D model $\bv$.
This 3D model is then \emph{decoded} into an estimate
\begin{equation}\label{e:rend}
\hat \x_0 (\x_t)
=
\Psi(\bv,\Pi)
= \Psi(\Phi(\x_t,\Pi,\sigma_t),\Pi)
\end{equation}
of the \emph{clean viewset} by the decoder $\Psi$ that implements differentiable rendering.
This is the same formulation as standard image-based diffusion, except that (1) one generates a set of views in parallel instead of a single image and (2) the denoiser network has a particular structure and geometric interpretation.
The training loss is the same as for standard diffusion:
$$
\mathcal{L}(\Phi, \x_0, \x_t, \Pi,t)
=
w(\sigma_t) \left \|
\Psi(\Phi(\x_t,\Pi,\sigma_t),\Pi)
- \x_0
\right \|^2
$$
where $\x_t = \sqrt{1- \sigma_t^2}\,\x_0 + \sigma_{t} \beps_t$ is a noised version of the (clean) input viewset $\x_0$.

\paragraph{Single and few-view reconstruction.}

With the model above, we can learn \emph{simultaneously} unconditional 3D generation as well as single and few-view reconstruction with almost no changes.
Given a conditioning viewset $(\y,\Pi')$, in fact, we can sample $p(\x|\Pi,\y,\Pi')$ by feeding into the network $\Phi$ a mixture of noised and clean views:
$$
\bv
= \Phi(
  \x_t \oplus \y,
  \Pi \oplus \Pi',
  \sigma_t \oplus \mathbf{0}
)
$$
where $\oplus$ denotes concatenation along the view dimension.
Here $\sigma_t \oplus \mathbf{0}$ means that we treat $\sigma_t$ as a vector of noise variances, one for each view in the viewset, and append zeros to denote the fact that the conditioning views $\y$ are ``clean''.

\paragraph{Discussion.}\label{s:method_discussion}

The approach above learns a distribution $p(\x)$ over viewsets rather than a distribution $p(\bv)$ over 3D models.
As noted in \cref{s:introduction}, however, viewsets and 3D models can be thought to be in one-to-one correspondence, so sampling one is equivalent to sampling the other.
While this statement is correct in the limit of infinitely-large viewsets,\footnote{And ignoring inconsequential reconstruction ambiguities} crucially reconstruction in our case is performed by a network $\Phi$.
The benefit is that this reconstruction network can learn a 3D data prior and use it to perform 3D reconstruction with much greater data efficiency.
In fact, we use as few as 3 images per viewset, which are far from sufficient to optimise a radiance field from scratch.

Our approach is also related to RenderDiffusion (RD)~\cite{anciukevicius2022renderdiffusion}, but with substantial theoretical and practical differences.
First, using our notation, their approach amounts to reducing the size of the viewset to a \emph{single view}, which is insufficient to adequately represent a 3D object $\bv$.
In our case, by using a non-trivial viewset, the generation of successive denoised samples ensures coherent and plausible appearance and shape from \emph{all} viewpoints, which is not guaranteed in RD, which only denoises a \emph{single} viewpoint.
We also introduce architectural advancements in the form of local conditioning and multi-view attention-based aggregation, further improving quality.

\subsection{Radiance fields and network architecture}\label{s:method_network}

\begin{figure}
\centering
\includegraphics[width=\columnwidth]{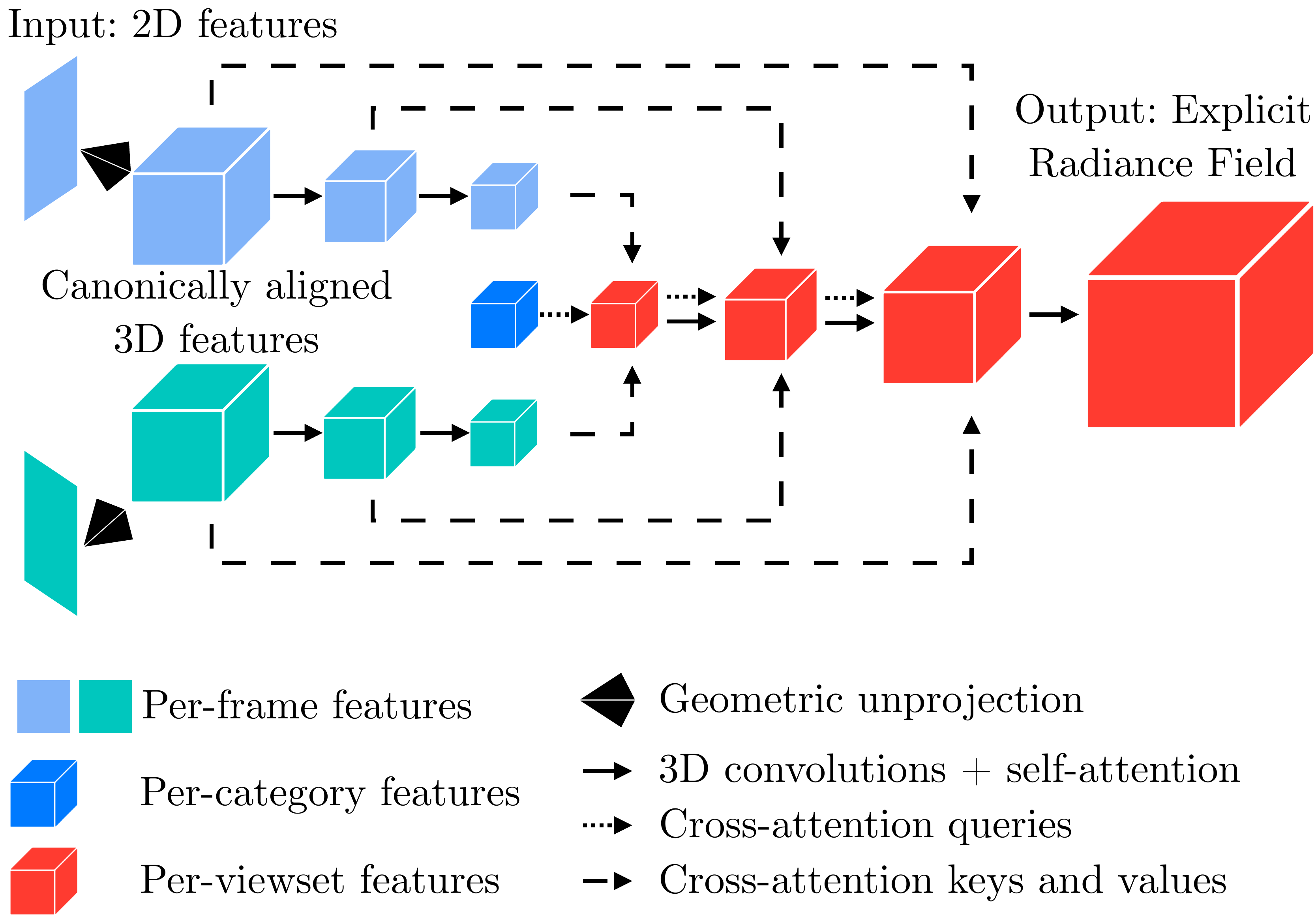}
\caption{\textbf{Architecture.}
2D input views are unprojected along camera rays to a canonical feature volume.
Multi-scale features are extracted and aggregated with an attention mechanism to output a single radiance field.
The number of input views can be variable.}%
\label{fig:architecture}
\end{figure}

Having discussed the learning formulation, we now describe key implementation details.
A 3D model $\bv=(\rho,\bc)$ is a \emph{radiance field}, i.e., a pair of functions $\rho$ and $\bc$ mapping a 3D point $q\in\mathbb{R}^3$ to an opacity value $\rho(q) \geq 0$ and an RGB color $\bc(q) \in [0,1]^3$.
For simplicity, we discretize the radiance field over a 3D grid, expressing it as a tensor $\bv \in \mathbb{R}^{4\times H\times W\times D}$, and evaluate $(\rho(q),\bc(q))$ using trilinear interpolation and padding as needed.
Similar to DVGO~\cite{sun22direct}, the voxel grid stores colors and opacities pre-activation, and activations are applied only after trilinear interpolation.
Given the camera $\pi$ (specified by rotation, translation and intrinsic parameters such as the focal length), then \cref{e:rend} \emph{renders} the image $x = \Psi(\bv,\pi)$ in a differentiable manner via ray casting.

The goal of the network $\Phi$ of \cref{e:encoder} is to output the 3D model $\bv$ given as input the viewset $\x_t$ (including the cameras $\Pi$) and the noise variance $\sigma_t$ for each view.
Our network consists of the following stages (see~\cref{fig:architecture}):

\paragraph{1. 2D feature extraction.}

A small 5-layer 2D convolutional subnetwork $f$ outputs a feature map $F^{(i)} = f(x_t^{(i)})$ for each image $x_{t}^{(i)}$ in out of $N$ images in viewset $\x_t$.

\paragraph{2. Geometric unprojection.}

For each feature map $F^{(i)}$, associated camera pose $\pi^{(i)} = (R^{(i)}, T^{(i)})$ and the camera intrinsic matrix $K$, we form a volume $V^{(i)} \in \mathbb{R}^{C \times H \times W \times D}$.
Voxel with centre at location $q = (i, j, k)$ holds feature $ F^{(i)}\left[ K \left( R^{(i)} | T^{(i)} \right)~ \tilde{q} \right]$, where $\left[ \cdot \right] $ denotes bilinear interpolation, $\Tilde{\cdot}$ denotes homogeneous coordinates and $(\cdot|\cdot)$ denotes column-wise matrix concatenation.
After this step, volumes $V^{(i)}$ for different images $x_t^{(i)}$ share the same global reference frame, so they ``line up''.
Unprojection allows the volume to easily match the conditioning image.

\paragraph{3. Per-frame 3D U-Net encoder.}

We use the same U-Net encoder as in DDPM~\cite{ho20denoising}, but replace the 2D convolutional and self-attention blocks with their 3D equivalents, similarly to~\cite{mueller2022diffrf}.
The encoder outputs multi-scale feature maps $\{W_{j}^{(i)} \}_{j=1}^{M}$, with $j=1$ being the finest and $j=M$ the coarsest feature map, for each input volume $V^{(i)}$.
We also pass the timestep $t$ (and thus, implicitly, the noise level $\sigma_t$) to the encoder after via the Transformer sinusoidal positional embedding~\cite{vaswani17attention}.
Similarly to DDPM~\cite{ho20denoising}, the timestep modulates the U-Net Convolutional blocks via FiLM~\cite{dumoulin18feature}.
Each input volume $V^{(i)}$ is processed independently by the encoder, hence it accepts the individual noise level $\sigma^{(i)}$ for corresponding image $I^{(i)}$.

\paragraph{4. Multi-view 3D U-Net decoder.}

The decoder acts as a multi-scale multi-view feature aggregator. At each level $j$ of the U-Net, the decoder aggregates features the feature maps $\{ W_{j}^{(i)} \}_{i=1}^{N}$ at level $j$ with an attention mechanism: $W_{j-1}' = \operatorname{Attn} (\text{Q} = Q_{j}, \text{K} = \text{V} = \{W_{j}^{(i)}\}_{i=1}^{N})$. The query $Q_{M}$ at the coarsest level is fixed and learnt per-class.
Attention operates at each voxel location independently to minimise computational complexity.
Learnt attention-based aggregation (instead of mean-averaging) means that the combination of features across views can depend on, for example, occlusion.
At each feature map level $j$, the aggregated features $W_{j-1}'$ are then upscaled, passed through convolutional and self-attention blocks, identically to usual U-Nets used in diffusion models~\cite{ho20denoising} to output $Q_{j-1} = h(W_{j-1}')$, before aggregation at the finer level $j-1$.

\paragraph{5. Upscaling.}

Finally, a small 5-layer 3D convolutional subnetwork $g$ performs upscaling $\bv = g(Q_0)$ to output the reconstructed volume $\bv$.
The output of the network is a single volume $\bv$ for the $N$ input views in viewset $\x_t$.

We validate our design choices, including the use of unprojection and attention-based aggregation in \cref{tab:ablation}.

\subsection{Training and inference details}%
\label{s:training_details}

\paragraph{Training.} To train our model, we consider a dataset $\mathcal{X}$ of viewsets (for example, from CO3D).
We further subsample each viewset extracting at random $N_\text{train}$ views $(\x, \Pi)$ that will be passed at input to the network, where $N_\text{train} \in {1, 2}$, so that $\x \in \{ \{ x^{(1)}\}, \{x^{(1)}, x^{(2)}\}\}$, and an additional unseen view $(x_u, \pi_u)$, which is unavailable to the network.
We sample the noise level $t$, and with it the scalar noise variance $\bar\sigma_t$ according to the cosine schedule of~\cite{nichol2021}.
We then randomly apply Gaussian noise to some of the input views, such that noise standard deviations can be in one of three states
$\sigma_t \in \{ \{  \bar\sigma_t \}, \{\bar\sigma_t, \bar\sigma_t\}, \{\bar\sigma_t, 0\} \}$, corresponding to one noised view, two noised views, or one clean and one noised view, respectively.
These three options are sampled with probability $[0.45, 0.45, 0.1]$, respectively.
The noised viewset $\x_t = \alpha_t \x + \sigma_t \beps_t$ is input to the network\footnote{With a slight abuse of notation, this applies a given noise level to the corresponding image.
Also note that some elements of $\sigma_t$ can be $0$, therefore $\x_t$ can also contain clean (non-noised) images.}.

\paragraph{Loss.}

We optimise the network $\Phi$ by minimising the photometric $L2$ loss (\cref{s:method_viewset}), of the renders from reconstructed volume:
\begin{multline}
\mathcal{L} = \sum_{i=1}^{N} w(\sigma^{(i)})\,
\| \Psi (\Phi (\x_t, \Pi, \sigma_t), \pi^{(i)}) - x^{(i)} \|^{2} \\ +
\lambda_t
\| \Psi (\Phi (\x_t, \Pi, \sigma_t), \pi_u) - x_{u} \|^{2}.
\end{multline}
The weights $w(\sigma^{(i)})$ are set according to the Min-SNR-5 strategy~\cite{hang2023minsnr}.
The loss $\mathcal{L}$ also includes a penalty on the unseen view $(x_u, \pi_u)$, as a regularisation strategy to encourage 3D consistency.
The weight of the unseen view $\lambda_t = \lambda \cdot \min_{i}w(\sigma^{(i)})$ is also noise-dependent, where $\lambda$ is a hyperparameter.

\paragraph{Inference.}

Inference is performed by progressive denoising of the viewset $\x$.
The size of the viewset $N_\text{inf}$ used at inference depends on the dataset and its complexity --- we use $N_\text{inf}=5,~3,~4$ for CO3D, ShapeNet and Minens, respectively.
When performing single-view reconstruction, the viewset $\x$ includes the clean conditioning view, which is unchanged during denoising the viewset.
In unconditional 3D generation, images in the viewset $\x$ are initialised to samples from Gaussian distribution.
We use DDIM~\cite{song2020denoising} sampling with 250 steps.
\section{Experiments}\label{s:experiments}

\paragraph{Datasets.}

We evaluate our method at $128 \times 128$ resolution on \emph{ShapeNet-SRN Cars}~\cite{sitzmann2019srns} using the standard train/val/test split~\cite{sitzmann2019srns} and protocol:
view 64 is used for conditioning and the remaining 250 views as novel, unseen prediction targets.
We also evaluate our method on four object classes from \emph{CO3D}~\cite{reizenstein21co3d}: Hydrant, Teddybear, Plant and Vase.
For each object class, we form a small testing set with 100 examples of randomly sampled image pairs and associated camera poses, one for conditioning and the other as the target, from randomly sampled test instances.
Pre-processing details for CO3D are in the sup.~mat.

\begin{figure}
\centering
\includegraphics[width=0.95\columnwidth]{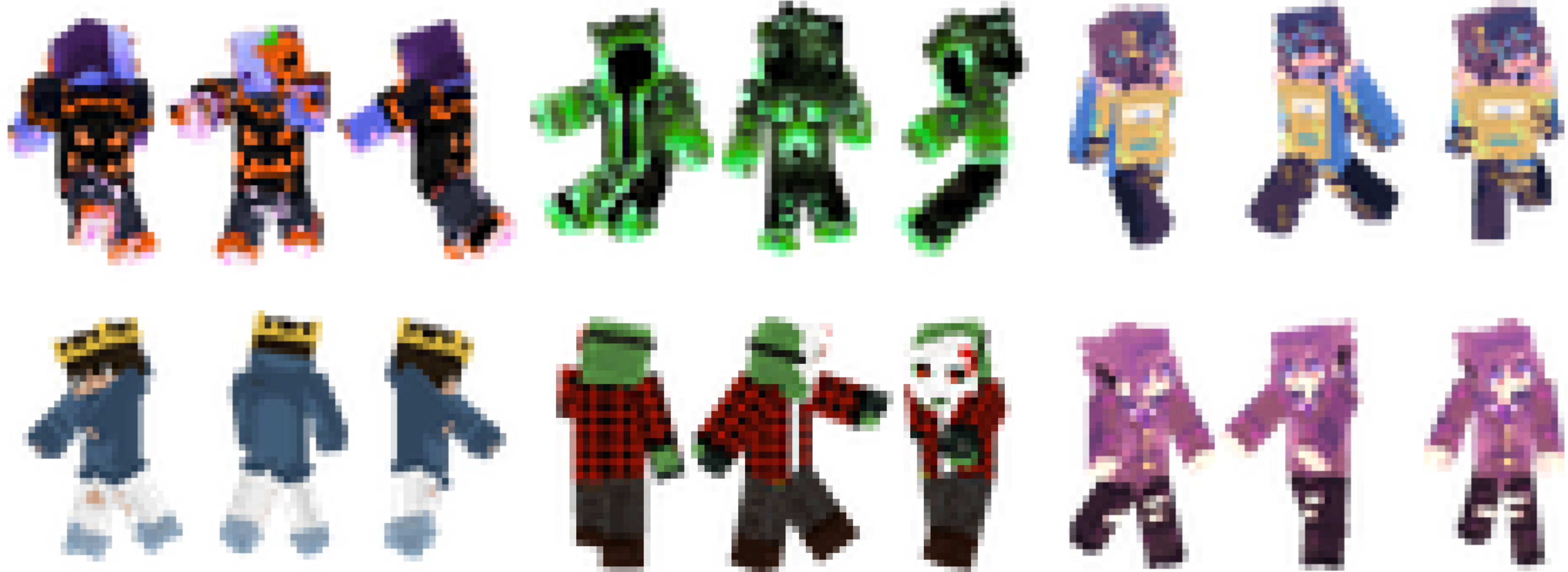}
\caption{\textbf{Minens dataset.}
Textured meshes are articulated and rendered from random camera viewpoints, allowing for procedural generation of a large number of instances.}\label{fig:minecraft_data}
\end{figure}

We also introduce \emph{Minens} (see \cref{fig:minecraft_data}), a new dataset that makes it easier to evaluate ambiguity and diversity in 3D reconstruction.
We design it to be large in the number of instances, while sufficiently small for rigorous experimentation using academic resources.
Each object in Minens consists of a torso with randomly articulated arms, legs and head and is textured with one of 3,000 skins.
We render 40,000 training and 5,000 validation examples with OpenGL at 256$\times$256 resolution and downsample to 48$\times$48 with Lanczos interpolation.
Each example consists of 3 images and associated camera poses $\Pi$.
We form two test sets with the Minens dataset:
`Random', with randomly sampled skins, poses and camera viewpoints, and `Ambiguous', with manually-selected 3D poses that are ambiguous when seen from a single viewpoint, \eg, due to one arm being occluded by the torso.
We use different skins for training and testing.
Similar to our subset of CO3D above, we select 100 test samples, consisting of one conditioning image and one target image, from different viewpoints.
Code and the Minens dataset are available at \url{szymanowiczs.github.io/viewset-diffusion}.

\paragraph{Evaluation protocol.}

We render the reconstructed object from the target viewpoint(s) and measure the Peak Signal-to-Noise Ratio (PSNR), Structural Similarity Index (SSIM) and Learned Perceptual Image Patch Similarity~\cite{zhang2018perceptual} (LPIPS), measured with a pre-trained VGG Net~\cite{simonyan15very}, when compared to the ground truth image.
However, deterministic baselines like PixelNeRF produce a single average reconstruction which is optimal for squared-error metrics like PSNR\@:
the `average' sample is blurry (\cref{fig:ambiguities}), but closer in PSNR to the ground truth than most samples taken from the correct posterior distribution.\footnote{%
By definition, the mean of a distribution has the minimum average square distance to all samples, and this distance is significantly less than the average squared distance between pairs of samples, particularly in high-dimensional spaces like images.}
Hence, by only looking at metrics like PSNR, it is simply not possible to measure the benefits of modelling ambiguity.
To measure the latter, we take multiple samples of reconstructions of every object and report the \emph{best} PSNR and SSIM from these samples.
For Minens and CO3D we use 100 samples per testing instance.
For ShapeNet-SRN, we take 20 samples due to the computational burden (the benchmark tests 175,000 generated views).
As LPIPS is not as strongly affected by this property since it measures perceived visual similarity, here we report the \emph{average} across all 100/20 samples.
For completeness, we also report the best LPIPS across all samples and the average PSNR and SSIM, but we report them in brackets `$(\cdot)$' as they do not measure well what we want.

\paragraph{Baselines.}

Our primary aim is to show the importance of modelling reconstruction ambiguity by showing that this results in sharper and possibly more accurate reconstructions than deterministic predictors.
We compare against other reconstructors trained using 2D data:
the fully-deterministic PixelNeRF~\cite{yu21pixelnerf:},
our reimplementation of RenderDiffusion (RD)~\cite{anciukevicius2022renderdiffusion},
and our improvement over it, RD++.
We train PixelNeRF using the publicly available code with a tuned learning rate and softplus activation for improved training stability and
reimplement RD based on publicly available information.
Since RD uses only single images for training (using our notation from \cref{s:training_details}: $N_\text{train}=1$, $N_\text{inf}=1$, $\sigma_t = \{ \bar\sigma_t \}$ and $\lambda = 0$) and a weaker architecture, for fairness we also consider the variant RD++.
RD++ uses the architecture from~\cref{s:method_network} (like Viewset Diffusion), takes as input a single noised image (like RD) and is trained with multi-view supervision (like Viewset Diffusion):
$N_\text{train}=1$, $\sigma_t = \{ \bar\sigma_t \}$ and $\lambda \neq 0$.
At inference time RD++, like RD, is capable of 3D generation by diffusing over a single view $N_\text{inf}=1$, starting from pure Gaussian noise.
Like RD, RD++ performs single-view reconstruction in a deterministic manner by accepting one clean input view $N_\text{inf}=1$, $\sigma_t = \{ 0 \}$.
Finally, to evaluate the importance of a probabilistic model, we include a baseline of our method without diffusion (\ie, one that receives a clean image and directly regresses a 3D volume): $N_\text{train} \in \{1,2\}$, $\sigma_t \in \{\{ 0 \}, \{ 0, 0 \} \}$, $N_\text{inf}=1$.

On ShapeNet-SRN we compare to single-view reconstruction deterministic works which report scores on this standard benchmark~\cite{guo2022fenvs,jang21codenerf:,lin2023visionnerf,sitzmann2021lfns,sitzmann2019srns,watson20223dim,yu21pixelnerf:}.

\subsection{Single-view reconstruction on Minens}%
\label{s:experiments_minecraft}

\begin{table}
  \centering
  \resizebox{0.47\textwidth}{!}{%
  \begin{tabular}{l l l l l }
  \toprule
    Method & \multicolumn{2}{c}{Random} & \multicolumn{2}{c}{Ambiguous} \\
    {} & PSNR $\uparrow$ & LPIPS $\downarrow$ & PSNR $\uparrow$ & LPIPS $\downarrow$ \\
    \toprule
    RenderDiffusion & ~19.85 & ~0.213 & ~16.33 & ~0.236 \\
    PixelNeRF & ~21.55 & ~0.220 & ~17.86 & ~0.250 \\
    RenderDiffusion++ & ~24.18 & ~0.157 & ~19.92 & ~0.210 \\
    Ours w/o $\mathcal{D}$ & ~24.63 & ~0.115 & ~20.26 & ~0.156 \\
    \midrule
    Ours w $\mathcal{D}$ - best & ~\textbf{24.82} & (0.072) & ~\textbf{21.50} & (0.081) \\
    Ours w $\mathcal{D}$ - mean & (22.81) & ~\textbf{0.107} & (18.62) & ~\textbf{0.130} \\
    \bottomrule \\
  \end{tabular}
  }
  \caption{\textbf{Single view reconstruction - Minens.} Ours achieves larger gains in the Ambiguous subset, showcasing the strength of probabilistic modelling.}
  \label{tab:synthetic_single_view_reconstruction}
\end{table}

\begin{table}
  \centering
  \begin{tabular}{l l l l }
  \toprule
    Method & PSNR $\uparrow$ & SSIM $\uparrow$ & LPIPS $\downarrow$ \\
    \toprule
    3DiM & ~21.01 & ~0.57 & - \\
    LFN & ~22.42 & ~0.89 & - \\
    \midrule
    SRN & ~22.25 & ~0.88 & ~0.129 \\
    CodeNeRF & ~22.73 & ~0.89 & ~0.128 \\
    FE-NVS & ~22.83 & (0.91*) & (0.099*) \\
    VisionNeRF & ~22.88 & ~\underline{0.90} & ~\textbf{0.084} \\
    PixelNeRF & ~23.17 & ~0.89 & ~0.146 \\
    Ours w/o $\mathcal{D}$ & ~\underline{23.21} & ~\underline{0.90} & ~0.116 \\
    \midrule
    Ours w $\mathcal{D}$ - best & ~\textbf{23.29} & ~\textbf{0.91} & (0.094) \\
    Ours w $\mathcal{D}$ - mean & (22.72) & (0.90) & ~\underline{0.099} \\
    \bottomrule \\
  \end{tabular}
  \caption{\textbf{Single view reconstruction --- ShapeNet Cars.} Ours achieves the best PSNR, with the additional benefit of probabilistic treatment. *FE-NVS optimises SSIM in training, affecting perceptual sharpness.}%
  \label{tab:single_view_reconstruction_shapenet}
  \vspace{-0.3cm}
\end{table}

In \cref{tab:synthetic_single_view_reconstruction} we compare the reconstruction quality using PSNR and LPIPS\@.
Sampling multiple plausible reconstructions via views diffusion improves  the PSNR of the best sample in both `Ambiguous' and `Random' subsets.
The significant gain in PSNR in the `Ambiguous' dataset shows that diffusion can effectively single-view reconstruction ambiguities.
Renders of samples of the reconstructed volumes in \cref{fig:ambiguities} show how diverse poses and textures are sampled under the presence of ambiguity.

\begin{figure}
\centering
\includegraphics[width=\columnwidth]{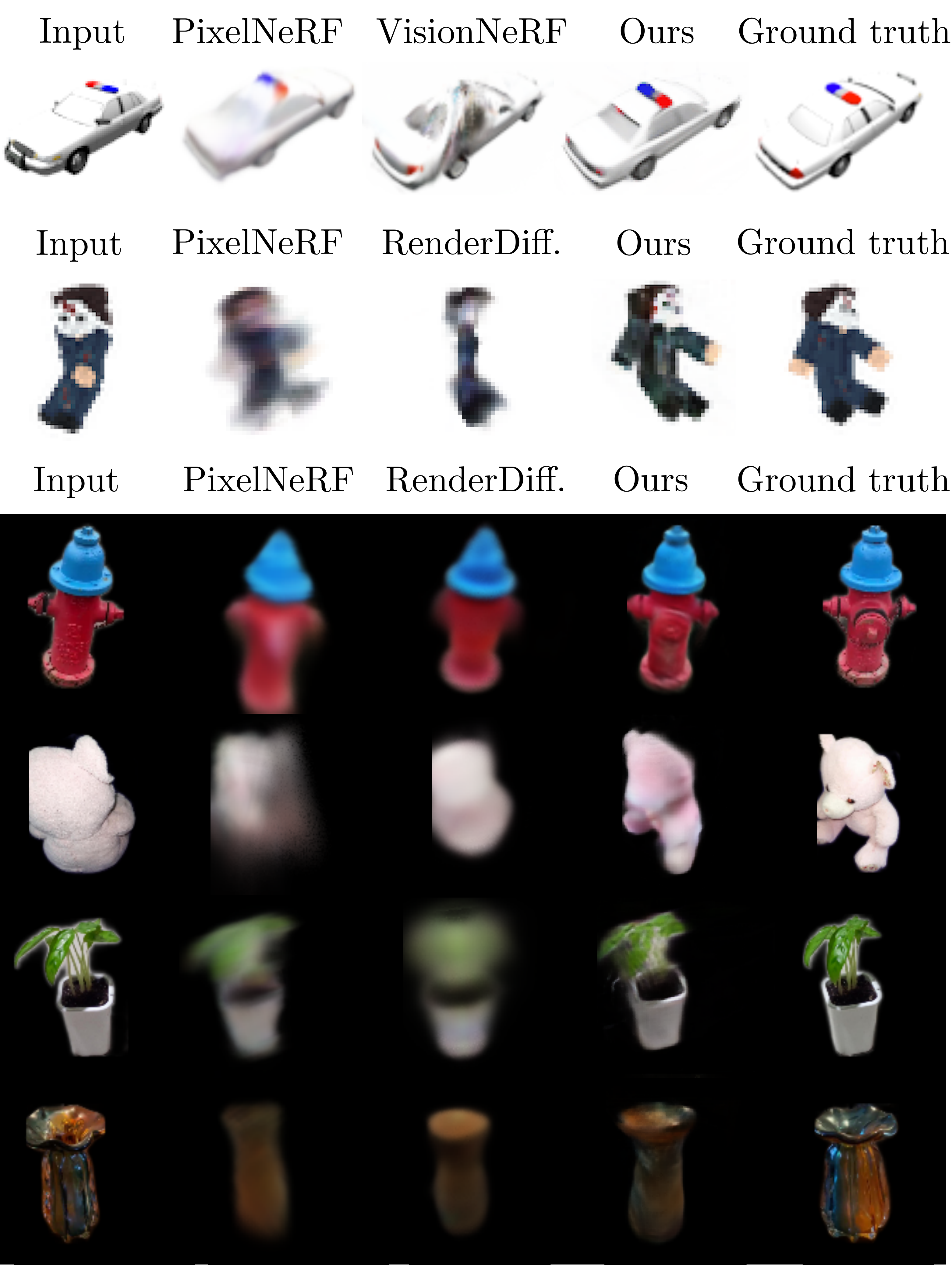}
\caption{\textbf{Single view reconstruction}. Our method outputs sharper shapes than prior work. The solutions are ambiguous, therefore our samples do not match the ground truth exactly but are more plausible than deterministic baselines.}
\vspace{-0.4cm}%
\label{fig:reconstruction}
\end{figure}

In \cref{tab:synthetic_single_view_reconstruction} it is also seen that using views diffusion leads to a decrease in \emph{average} LPIPS (lower is better), suggesting that \emph{all} samples from our method are more perceptually plausible than the results from the baselines.
Finally, the improvement in the metrics is further accompanied by qualitative comparison in \cref{fig:ambiguities,fig:reconstruction} where our samples are seen to be much sharper than the baseline results.

\subsection{Single-view reconstruction: ShapeNet \& CO3D}

Quantitative results in ShapeNet and CO3D are given in~\cref{tab:single_view_reconstruction_shapenet,tab:single_view_co3d}.
Like in Minens, in ShapeNet-SRN the best sample from Viewset Diffusion has higher PSNR than the baselines, which indicates that the model can sample from the correct distribution.
This is further confirmed by Viewset Diffusion's lower (better) LPIPS than all baselines in CO3D (\cref{tab:single_view_co3d}) and almost all baselines in ShapeNet (\cref{tab:single_view_reconstruction_shapenet}).
VisionNeRF~\cite{lin2023visionnerf} outperforms our method in LPIPS, likely due to their use of the much stronger ViT-based 2D feature extraction (our 2D feature extractor is much smaller, consisting of only 5 convolutional layers).
On challenging CO3D classes (Teddybear, Plant, Vase), the deterministic baselines achieve better PSNR, possibly because more than 100 samples are required to adequately sample the space of reconstructions for these more complex objects.

\begin{table*}
  \centering
  \resizebox{0.97\textwidth}{!}{%
  \begin{tabular}{l l l l l l l l l l l l l }
    \toprule
    {} &
    \multicolumn{3}{c}{Hydrant} &
    \multicolumn{3}{c}{Teddybear} &
    \multicolumn{3}{c}{Plant} &
    \multicolumn{3}{c}{Vase} \\
    \cmidrule(lr){2-4}\cmidrule(lr){5-7}\cmidrule(lr){8-10}\cmidrule(lr){11-13}
    Method & PSNR $\uparrow$ & SSIM $\uparrow$ & LPIPS $\downarrow$ &
    PSNR $\uparrow$ & SSIM $\uparrow$ & LPIPS $\downarrow$ &
    PSNR $\uparrow$ & SSIM $\uparrow$ & LPIPS $\downarrow$ &
    PSNR $\uparrow$ & SSIM $\uparrow$ & LPIPS $\downarrow$ \\
    \toprule
    RenderDiffusion & ~17.43 & ~0.70 &  ~0.263 &
    ~14.71 & ~0.48 & ~0.444 &
    ~17.30 & ~0.46 & ~0.467 &
    ~18.92 & ~0.68 & ~0.288 \\
    PixelNeRF & ~18.07 & ~0.67 & ~0.297 &
    ~15.01 & ~0.43 & ~0.451 &
    ~17.62 & ~0.41 & ~0.460 &
    ~17.98 & ~0.61 & ~0.329 \\
    RenderDiffusion++ & ~21.61 & ~0.69 & ~0.282 &
    ~19.58 & ~0.65 & ~0.303 &
    ~19.85 & ~0.49 & ~0.399 &
    ~21.51 & ~0.65 & ~0.292 \\
    Ours w/o $\mathcal{D}$ & ~22.06 & ~0.78 & ~0.217 &
    ~\textbf{19.73} & ~0.65 & ~\textbf{0.309} &
    ~\textbf{20.33} & ~0.51 & ~0.382 &
    ~\textbf{21.89} & ~0.68 & ~0.264 \\
    \midrule
    Ours w $\mathcal{D}$ - best & ~\textbf{22.36} & ~\textbf{0.80} & (0.176) &
    ~19.68 & ~\textbf{0.70} & (0.267) &
    ~20.23 & ~\textbf{0.58} & (0.339) &
    ~21.36 & ~\textbf{0.75} & (0.210) \\
    Ours w $\mathcal{D}$ - mean & (20.52) & (0.77) & ~\textbf{0.199} &
    (17.28) & (0.64) & ~\textbf{0.309} &
    (19.47) & (0.40) & ~\textbf{0.366} &
    (20.05) & (0.71) & ~\textbf{0.237} \\
    \bottomrule \\
  \end{tabular}
  }
  \caption{\textbf{Single view reconstruction - CO3D.} Our method improves over baselines in CO3D classes on SSIM and LPIPS.}\label{tab:single_view_co3d}
\end{table*}

\begin{figure*}
\centering
\includegraphics[width=\textwidth]{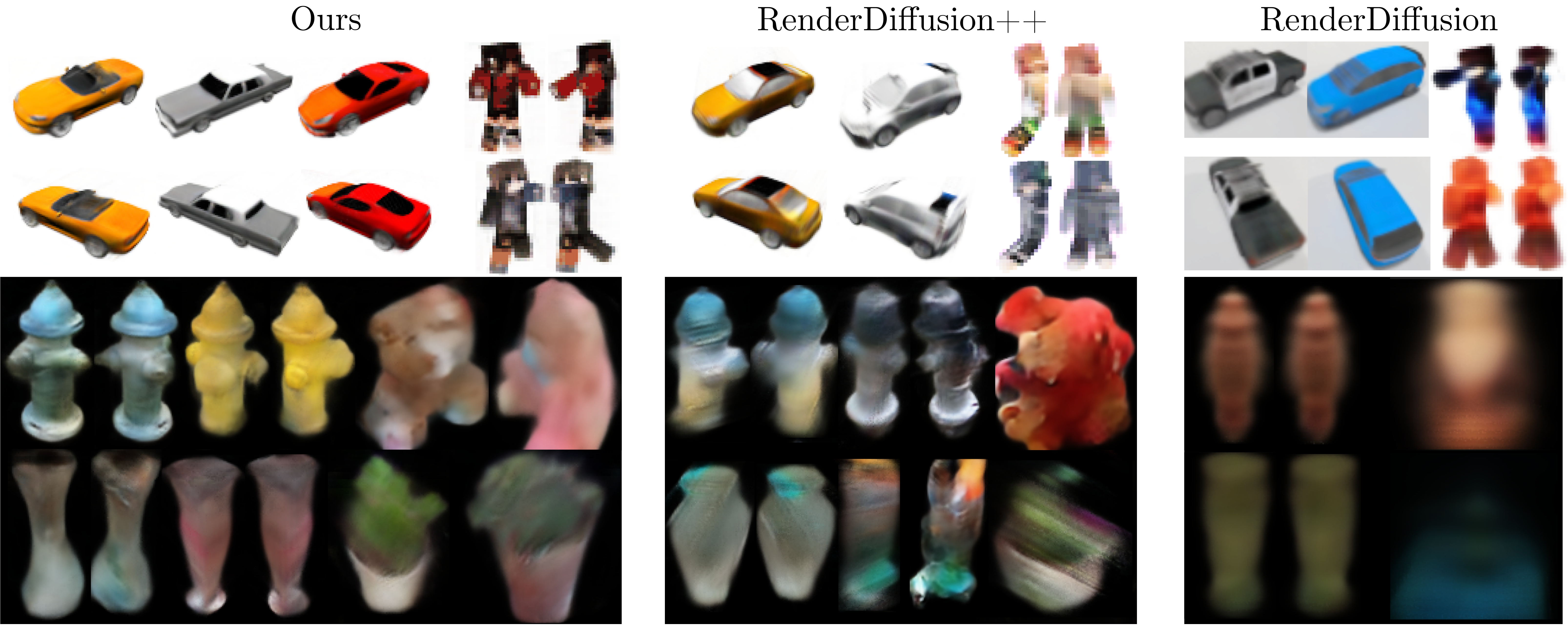}
\caption{\textbf{Unconditional generation - Cars, Minens, Hydrants, Teddybears, Vases, Plants.} Samples from our method show higher visual detail than RenderDiffusion~\cite{anciukevicius2022renderdiffusion} and our improvement over it, RenderDiffusion++.
}%
\label{fig:generation_hydrants}
\end{figure*}

\subsection{Unconditional generation}

Viewset Diffusion supports unconditional generation by setting the number of clean input views to $0$.
To generate 3D prior works~\cite{mueller2022diffrf, shue2022nfd, wang2023rodin} require 3D ground truth at training time, while we only use 3 views per object\footnote{For Minens dataset this is precisely true. In CO3D and ShapeNet different viewsets may come from the same objects due to data limitations, but they are treated independently by the training algorithm.} (at test time, we can generate any number of 3D consistent views).
In \cref{fig:generation_hydrants} we compare samples from our method (network from~\cref{s:method_network}, $N_\text{inf} > 1$), RD++ (network from~\cref{s:method_network}, $N_\text{inf} = 1$) and RD (network from~\cite{anciukevicius2022renderdiffusion}, $N_\text{inf} = 1$).
Viewset Diffusion samples are sharper from all viewpoints, which demonstrates the advantage of diffusing more than one view jointly.
Intuitively, in the last step of diffusion, Viewset Diffusion essentially performs reconstruction from $3 \leq N_\text{inf} \leq 5$ (nearly) clean views of the object, whereas RD and RD++ do so from a single view, which results in blurry images from other viewpoints.

\subsection{Ablations}%
\label{s:experiments_ablation}

We assess the importance of different components of our method: input image unprojection, attention-based aggregation of features from different views, and using diffusion ($\mathcal{D}$).
We use the `Ambiguous' Minens dataset and evaluate best PSNR and average LPIPS in the unseen novel views.
We train smaller models (half the number of U-Net convolutional layers and no self-attention layers) for fewer iterations (60k) due to the computational cost.
Results are reported in \cref{tab:ablation}.
Not using diffusion ($\sigma_t = \{ 0 \}$, using notation from \cref{s:training_details}) leads to a drop in PSNR and worse perceptual quality due to the reconstructions being blurry in the presence of ambiguity.
Removing attention-based feature aggregation (\cref{s:method_network}, 4.) across frames and aggregating them with a simple mean prohibits the network from reasoning about viewpoints and occlusion when pooling the features from different views.
Finally, removing unprojection (\cref{s:method_network}, 2.) hinders the learning process due to the removal of local conditioning which is known to improve the learning process~\cite{yu21pixelnerf:}.

\begin{table}
\centering
\begin{tabular}{l c c}
& PSNR $\uparrow$ & LPIPS $\downarrow$ \\
\toprule
Full model & \textbf{20.36} & \textbf{0.075} \\
\midrule
$\circleddash$ diffusion $ \mathcal{D}$ & 18.85 & 0.101 \\
$\circleddash$ attention in aggregation & 19.54 & 0.100 \\
$\circleddash$ unprojection & 18.26 &  0.164 \\
\bottomrule \\
\end{tabular}
\caption{\textbf{Ablations.} Impact of removing component from our method on reconstruction quality.}%
\label{tab:ablation}
\end{table}

\section{Conclusions}\label{s:conclusions}

We have presented Viewset Diffusion, a method to learn a model for probabilistic single-view 3D reconstruction and generation.
By diffusing viewsets, we can learn a DDPM from multi-view 2D supervision and still learn to generate 3D objects, having only 3 views per object and no access to 3D ground truth.
Viewset Diffusion also unifies 3D reconstruction and generation, and enables feed-forward probabilistic 3D reconstruction with diffusion models.
We have shown empirically that a probabilistic approach to the single-view reconstruction problem leads to higher-quality results and less blurry solutions than deterministic alternatives.

\paragraph{Ethics.}
We use the ShapeNet and CO3D datasets in a manner compatible with their terms.
The images used in this research do \emph{not} contain personal information such as faces.
For further details on ethics, data protection, and copyright please see \url{https://www.robots.ox.ac.uk/~vedaldi/research/union/ethics.html}.

\paragraph*{Acknowledgements.}

S.~Szymanowicz is supported by an EPSRC Doctoral Training Partnerships (DTP) EP/R513295/1 and the Oxford-Ashton Scholarship.
A.~Vedaldi and C.~Rupprecht are supported by ERC-CoG UNION 101001212.
C.~Rupprecht is also supported by  VisualAI EP/T028572/1.
{\small\bibliographystyle{ieee_fullname}\bibliography{more,vedaldi_specific,vedaldi_general}}

\end{document}


\maketitle

\ificcvfinal\thispagestyle{empty}\fi

\section{Additional results}
 Visit the project website \url{https://szymanowiczs.github.io/viewset-diffusion.html} for more visualisations of non-cherry-picked results of single-view 3D reconstruction and unconditional 3D generation across all classes.

\section{Minens Dataset}

\begin{figure*}
    \centering
    \includegraphics[width=0.9\textwidth]{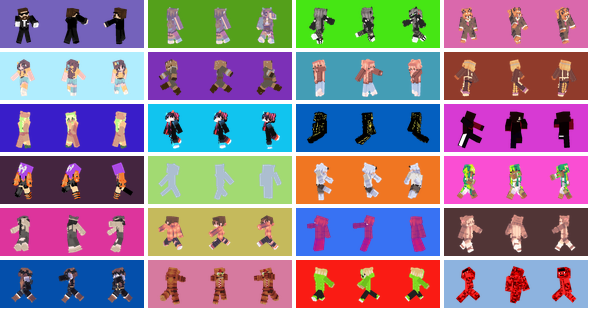}
    \caption{\textbf{Minens dataset -- training samples.} We show 24 random examples from the dataset. Each training example consists of 3 images (shown) and associated camera poses.}
    \label{fig:dataset_samples_general}
    
\end{figure*}

We provide further details on the Minens dataset we contribute as part of this work, in particular: distribution of camera poses, distribution of character articulation poses, image backgrounds, validation images and visualisations of both the training set and the `Ambiguous' test set.

\begin{figure}
    \centering
    \includegraphics[width=0.9\columnwidth]{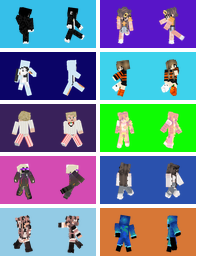}
    \caption{\textbf{Minens `Ambiguous' test set.} We show 10 random examples from the `Ambiguous' test set. In each example, left image is the input conditioning and right image is the ground truth testing view.}
    \label{fig:dataset_samples_ambiguous}
\end{figure}

\paragraph{Cameras.} Cameras are placed at a fixed distance from the origin, are directed towards the origin and have identical focal lengths and principal points.
Camera location is therefore parameterised by the azimuth and elevation angles that it forms with the world x-axis (the radius is fixed).
Yaw and elevation are sampled randomly and independently for each camera: azimuth is distributed uniformly in $[0, 2\pi]$ and elevation is distributed uniformly in $[-\frac{\pi}{8}; \frac{\pi}{8}]$.
\paragraph{Articulation.} Characters have a fixed torso location and orientation, facing in the direction of positive z-axis.
Both arms, both legs and the head are randomly and independently articulated.
Pitch of arms is distributed uniformly in $[-20^{\circ}, 45^{\circ}]$; roll is distributed uniformly in $[0^{\circ}, 10^{\circ}]$ for the left arm and $[-10^{\circ}, 0^{\circ}]$ for the right arm.
Pitch of legs is distributed uniformly in $[-30^{\circ}, 30^{\circ}]$; roll is distributed uniformly in $[0^{\circ}, 10^{\circ}]$ for the left leg and $[-10^{\circ}, 0^{\circ}]$ for the right leg.
Head pitch is distributed uniformly in $[-10^{\circ}, 10^{\circ}]$, head pith is distributed uniformly in $[-5^{\circ}, 5^{\circ}]$ and head yaw is distributed normally with mean $10^{\circ}$ and standard deviation $10^{\circ}$.
\paragraph{Backgrounds.} Character skins have varied colours, so there is no single background colour on which all skins clearly stand out from the background.
Therefore, for each example we choose a random background colour, with RBG distributed uniformly in $[0, 255]$.
The training examples are available with the alpha channel, therefore allowing for sampling of random backgrounds in every training iteration.
Testing examples are saved with a fixed, randomly chosen background channels so that metrics are consistently measured between different training runs.
Alpha channel was not used in any way other than to apply randomly coloured backgrounds, \ie we did not use masks to aid network training.
\paragraph{Validation viewpoints.} Each training example consists of 3 images and camera poses. 
In addition to the training images, each example (\ie each articulated character) is associated with a fourth image and camera pose which we do not use for training (\ie our network does not have access to it).
However, we can render the reconstructed character from the unseen viewpoint and we use the error in that viewpoint as the validation loss that approximates the quality of 3D shapes output by our method. 
\paragraph{Samples.}
\cref{fig:dataset_samples_general} shows random samples from the Minens training dataset: 3 images per character, rendered from different camera viewpoints. 
\cref{fig:dataset_samples_ambiguous} shows samples of test examples in the `Ambiguous' test set.
Ambiguities can be due to occlusion, where one limb is occluded by the torso.
Another type of ambiguity is projective ambiguity, \ie from a frontal image it can be ambiguous if a leg is in front of the body or behind the body.
Finally, there are ambiguities due to symmetry -- from a side image it can be ambiguous if the right leg is in front of the body and the left leg is behind, or if it is the other way around.
Samples in the `Ambiguous' test set exhibit more ambiguity than a randomly chosen test set would.
Our method shows the most improvement on the `Ambiguous' test set when compared to baselines.

\paragraph{Reproducing.}
We release the dataset in the form of code to run to exactly render the Minens dataset used for training, validation and testing. 

\section{Data}
\subsection{CO3D Preprocessing}
The data normalisation protocol for CO3D~\cite{reizenstein21common} objects aims to make objects approximately the same scale, place them in the centre of the voxel grid and align them vertically with the `world' vertical direction. 

\begin{figure}
    \centering
    \includegraphics[width=\columnwidth]{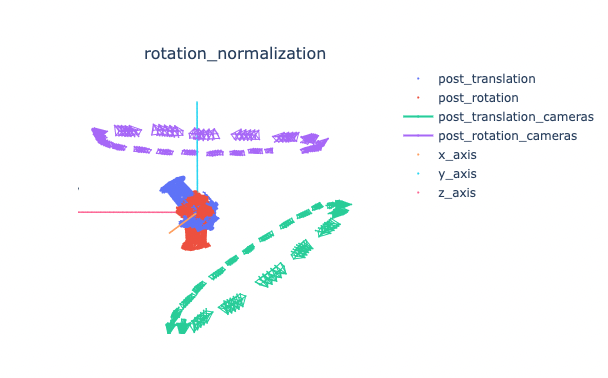}
    \caption{Rotation normalization. Red point cloud and purple cameras correspond to the translation and rotation aligned object. Blue points and green cameras correspond to the object after translation normalization, before rotation normalization.}
    \label{fig:rotation_normalization}
\end{figure}

\begin{enumerate}
    \item \textbf{Translation normalization.} 
    We find the centre of mass of the point cloud $\bar{x}$, shift the point cloud $\{ x^{(i)} \}$ so that its new centre of mass is at the world centre: $\{ x^{(i)} \}' = \{ x^{(i)} - \bar{x} \}$ and move the cameras accordingly: $T_{w2c}' = \bar{x} R_{w2c} + T_{w2c}$.
    \item \textbf{Rotation normalisation.} 
    We estimate the world 'up' direction by leveraging photographer's bias, \ie assuming that photos are taken with approximately zero yaw. Under this assumption, the camera x-vector is approximately perpendicular to the world direction. 
    We form a matrix by stacking x-vectors of all cameras in a sequence (normalized to $0$ mean) and run Singular Value Decomposition (SVD).
    $$ U \Sigma V^{T} = SVD ( [ 1, 0, 0 ] R_{w2c}^{T} ) $$ 
    $$ \hat{y} = V [ 0, 0, 1]^{T} $$
    SVD is only defined up to a direction ambiguity, therefore we set the world `up' vector and the camera vectors to point in the same direction, \ie ensure that $\hat{y}_{\text{world}} \cdot y_{\text{cam}} > 0$ and flip $\hat{y}$ if needed. 
    We use the the first camera for $y_{\text{cam}}$ but check that all cameras satisfy $y_{\text{cam1}} \cdot y_{\text{cam2}} > 0$ and exclude the sequence from training if that is not the case. 
    We also verify that that our assumption of the photographer's bias is valid in a sequence by checking that $\sigma^{2}_{1} / \sigma^{2}_{2} < \sigma^{2}_{2} / \sigma^{2}_3 $ and exclude a sequence if that is not the case.
    We find that most sequences pass these checks successfully therefore confirming our intuition about photographer's bias.
    A visualisation of rotation normalisation is shown in \cref{fig:rotation_normalization}.
    \item \textbf{Scale normalization.}
    To normalise scale we first shift the point cloud and cameras so that world centre is halfway between the top and bottom of the point cloud $\{ x^{(i)} \}'' = \{ x^{(i)} \}' - (y_{max} - y_{min})$. 
    This has effectively the same purpose as translation normalisation, but most videos are taken from above an object, so the point clouds are denser at the top of objects, meaning the point cloud mean $\bar{x}$ is nor an accurate object centre in the vertical direction.
    Finally, we want the scale of objects to be normalised across sequences, so we normalise the scene by the maximum absolute value of any point coordinate. The scaling factor $s$ for a voxel grid with side length $d$ is 
    $$s = \frac{d \times 0.95}{2 \times \text{max}_{i}|| x^{(i)} ||_{\infty}}.$$
    All point coordinates ${x^{(i)}}$ and camera locations are scaled by factor $s$.
\end{enumerate} 

We verify that after the normalisation the distances of the cameras are in sensible locations. 
In particular, for a volume of side length $1.2$ (which we use) we filter out sequences that have cameras very close to the centre, \ie with camera translation magnitude $||T||_{2} < 0.85$.
These sequences correspond to sequences with poor point cloud quality where some cameras were incorrectly estimated to be very close to the surface of the objects
We also filter out sequences with cameras very far from the centre, \ie with camera translation magnitude $||T||_{2} > 6.5$. 
Such sequences correspond to a failures in foreground / background segmentation which in turn lead to background being included in the point cloud.
As a result, scaling using maximum point location results in downscaling the scene too much.

Finally, some sequences in CO3D correspond to cameras being moved in front of a screen displaying objects instead of actual objects.
We filter out such sequences by imposing a minimum on standard deviation in depth values.
In total, filtering removes 99 sequences for the Hydrant class, 382 sequences for the Plant class, 293 sequences for the Vase class and 473 sequences for the Teddy bear class.
We do not remove any testing sequences.

We rescale all images to $128 \times 128$ resolution. As the focal lengths and principal points are provided in Normalised Device Coordinates, we do not need to rescale them when scaling the images or rescaling the world size.

\subsection{Pose encoding}

We encode the camera pose information in the tensor we pass to the network, similarly to 3DiM~\cite{watson20223dim}.
Each pixel RGB value is appended with an embedding of length $6$, holding the location of the camera origin in world coordinates (length $3$) and the normalised ray direction of the ray from the camera origin through the pixel (also length $3$). The pose encoding is appended along the channel dimension to the input RGB images.

\section{Baselines}
\subsection{PixelNeRF}

In PixelNeRF~\cite{yu21pixelnerf:} we use learning rate of $1e-5$ and use a batch size of $4$ for CO3D data (corresponding to 132 images, as each batch contains 32 images of the same object) and $32$ for Minens data (corresponding to 96 images as each batch contains 3 images of one object).
We use Softplus activation and tune the beta parameter that prevents collapse into 0 density region - $1.0$ for coarse network and $3.0$ for fine network. 
We train for single-view reconstruction for $100k$ iterations for the Minens dataset and $600k$ iterations for CO3D datasets. 
We use 64 coarse samples and $64$ fine samples at training time in the NeRF network for Minens dataset and $128$ coarse samples and $64$ fine samples for CO3D data.
We use the same normalised CO3D data for PixelNeRF and for our method.
For evaluation on ShapeNet-SRN we use pretrained PixelNeRF models with official evaluation code and metrics reported in the PixelNeRF paper~\cite{yu21pixelnerf:}.

\subsection{RenderDiffusion}

See~\cref{tab:baseline_architectures} for a summary of the baseline architectures, using notation from Sec.3.5 in the main paper.

\paragraph{RenderDiffusion re-implementation (RD).} We re-implemented RenderDiffusion with the publicly available information and we include the details and results here.
As in the original paper, we modified the U-Net~\cite{ho20denoising,ronneberger15u-net:} commonly used in diffusion models so that its output has $3n_{f}$ channels and reshaped it to form a triplane.
We used $n_{f}=32$ channels and a two-layer rendering MLP, with $32$ hidden units and an intermediate Softplus activation function.
Training was done for the same number of iterations as in our method with Adam~\cite{kingma15adam:} optimiser and the same hyperparameters as in our method.
The noise schedule used was the same as in our method.
We will release code for RenderDiffusion re-implementation together with our code.

\paragraph{RenderDiffusion++ (RD++).} Our initial experiments with the architecture that RenderDiffusion uses indicate that the textures that RD outputs are low frequency and lack high-frequency details present in the conditioning images, likely due to the absence of local conditioning.
Another issue we noticed was that the shapes output by RenderDiffusion are not necessarily plausible, \eg a reconstruction of a Minens character with three legs, possibly because the supervision in RenderDiffusion is single view.
While the reconstructions shown in the figures of RenderDiffusion have plausible shapes, they are only demonstrated for simple shapes from ShapeNet and CLEVR datasets with little ambiguity (\ie a single image is enough to reconstruct the 3D shape, \eg with symmetry constraints).
Our Minens dataset exhibits more ambiguity (\eg due to articulation) and thus the reconstructions output by RenderDiffusion are an average between the different plausible shapes.
Thus, for fair comparison, we also compare to RD++: RenderDiffusion, but with our architecture and with multi-view supervision.
We train a network that receives a single image of an object, with added Gaussian noise, and is tasked with predicting a reconstruction of the clean object.

\begin{table*}[]
    \centering
    \begin{tabular}{l c c c c c c c}
        \toprule
        {} & 3D & {} & {} & {} & {} & 3D & 3D \\
        Method & Representation & $\sigma_t$ & $N_{train}$ & $N_{inf}$ & $\lambda = 0 $ & Generation? & Reconstruction \\
        \midrule 
        RD & Triplane & $\{\bar\sigma_t\}$ & 1 & 1 & Yes & Yes & Deterministic \\
        RD++ & Grid & $\{\bar\sigma_t\}$ & 1 & 1 & No & Yes & Deterministic \\
        Ours w/o $\mathcal{D}$ & Grid & $\{0\}, \{0, 0\}$ & 1-2 & 1 & No & No & Deterministic \\
        Ours $\mathcal{D}$ & Grid & $\{\bar\sigma_t\}, \{\bar\sigma_t, 0\}, \{\bar\sigma_t, \bar\sigma_t\}$ & 1-2 & 3-5 & No & Yes & Generative \\
        \bottomrule \\
    \end{tabular}
    \caption{\textbf{Baseline methods.} We summarise key differences of baselines which we compare Viewset Diffusion (Ours w $\mathcal{D}$) against. The methods vary in number of images used at input in training $N_{train}$ and inference $N_{inf}$ and what levels of noise $\sigma_t$ are applied to the input images during training.
    In some baselines the unseen view is not penalised $\lambda=0$, only a subset of them is able to perform 3D Generation and only our method performs 3D Reconstruction in a generative manner.}
    \label{tab:baseline_architectures}
\end{table*}

\section{Technical details}

\subsection{Optimization.} 
We optimise the parameters of our network with Adam~\cite{kingma14adam:} optimizer and learning rate $2 \times 10^{-5}$. 
We use batch size 16 and optimise all diffusion networks for 100k (ShapeNet) / 200k (Minens, Hydrant, Teddybear) / 256k (Vase) / 280k (Vase) iterations and the networks without diffusion for 40k iterations. 
Timestep-dependent weighting strategy $w(\sigma^{(i)})$ is the Min-SNR-5 strategy~\cite{hang2023minsnr}: $w(\sigma^{(i)}) = \min \{ \text{SNR}^{(i)}, 5\}$.
SNR is the Signal-to-Noise Ratio: $\text{SNR}^{(i)} = (1-\sigma^{(i)2})/\sigma^{(i)2}$. 
Hyperparameter $\lambda$ for penalising unseen view is $\lambda=0.1$ in Minens and ShapeNet and $\lambda=0.2$ in CO3D.

\subsection{Diffusion.}
We use a diffusion schedule with 1000 diffusion steps and a cosine noise schedule. 
Our networks are trained with ``$x_{0}$'' formulation.
At inference, we use 250 steps of DDIM~\cite{song2020denoising} sampling.
Quantitative single-view reconstruction results cited in the main paper were obtained with $N-1$ noisy views and $1$ clean view in the viewset, where $N=3$ for ShapeNet-SRN, $N=4$ for Minens and $N=5$ for CO3D.
For generation we used a the same models and viewsets with the same size $N$, but without any clean views.
Varying numbers of images in the viewset are due to varying complexity of data and varying amount of ambiguity.

\subsection{Architecture}

\begin{figure*}
    \centering
    \includegraphics[width=0.9\textwidth]{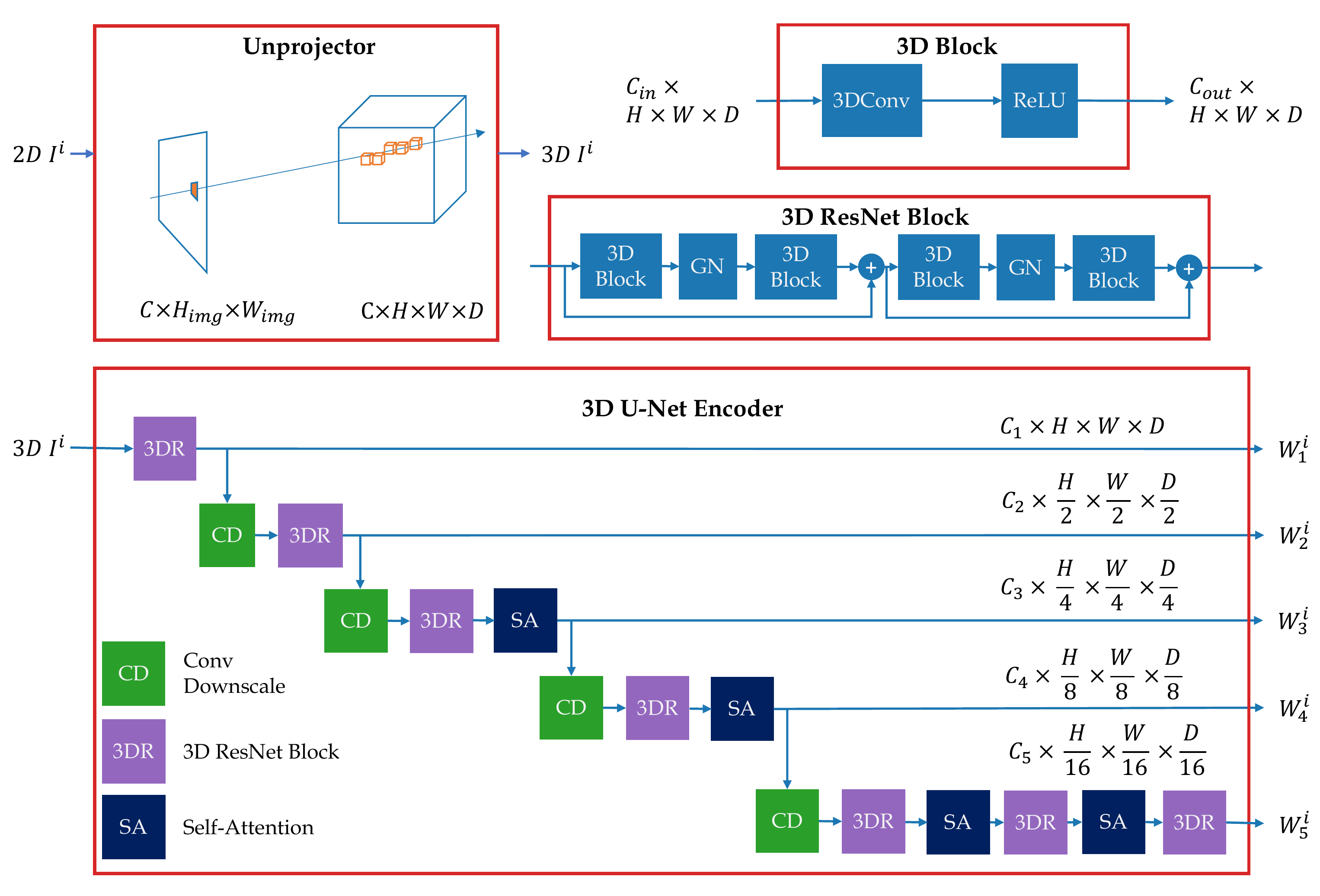}
    \caption{\textbf{Encoder.} Each image $I^{i}$ out of $N$ images input to the reconstructor is first unprojected to 3D (top left) and passed through a series of 3D ResNet Blocks (top right) and Max Pooling layers to output $M=5$ feature maps $W_{j}^{i}$.}
    \label{fig:architecture_encoder}
\end{figure*}

Our architecture takes in $N$ input frames, each of which can be clean or noisy, and outputs one volume of size $S \times S \times S$. 
In Minens $S=32$, in all other datasets $S=64$.
Each entry in the output volume holds $4$ values: $3$ for RGB colour and $1$ for opacity.
\paragraph{Encoder.} Each of the $N$ images is first passed through a small 2D CNN and an `Unprojector' (\cref{fig:architecture_encoder}, top left) which pools the RGB colours from the image into a 3D volume along camera rays. 
Next, each image is independently passed through an encoder which contains a series of 3D ResNet and MaxPool blocks (\cref{fig:architecture_encoder}, bottom). All convolutions are done with $3 \times 3 \times 3$ kernels. GroupNorm layers~\cite{wu18groupnorm} with $8$ groups are applied in every ResNet block to the intermediate output (\cref{fig:architecture_encoder}, top right). There are 4 Convolutional Downscaling Blocks in the encoder, outputting $M=5$ feature maps for each of $N$ images $I^{i}, \; i \in {1, \dots, N}$. We use channel dimensions $C_{1},C_{2},C_{3},C_{4},C_{5}=64,64,128,256,512$.

\begin{figure*}
    \centering
    \includegraphics[width=0.9\textwidth]{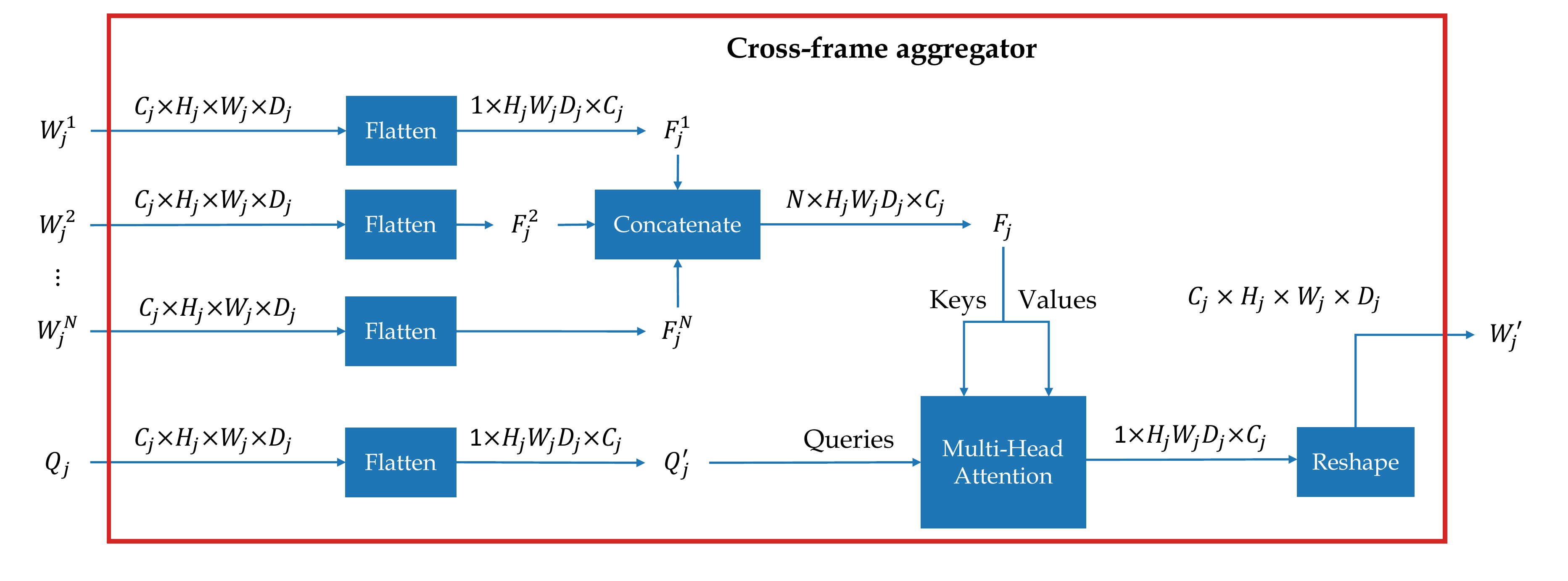}
    \caption{\textbf{Aggregator} takes in $N$ feature volumes $W_{j}^{i}$ at level $j$ as well as a query volume $Q_{j}$ and outputs an aggregated feature volume $W_{j}^{'}$.}
    \label{fig:architecture_aggregator}
\end{figure*}

\paragraph{Aggregator.} Given $N$ feature maps $W_{j}$ at level $j$ coming from $N$ images (\cref{fig:architecture_aggregator}, left), the `Aggregator' is tasked with aggregating the features into one feature volume $W_{j}'$ at level $j$ (\cref{fig:architecture_aggregator}, right). 
Each feature volume is flattened (\cref{fig:architecture_aggregator}, top left) and the flattened vectors are concatenated along the sequence dimension, forming a tensor of size $N \times H_{j}W_{j}D_{j} \times C_{j}$, which is input to an Attention layer with a GroupNorm layer and a residual skip-connection (\cref{fig:architecture_aggregator}, bottom middle). 
Concatenated features $F_{j}$ are input as the Keys and Values to the attention layer, with the number of frames $N$ being the sequence dimension and the flattened voxels being the batch dimension.
Query volume $Q_{j}$ for feature level $j$ has size $1 \times H_{j}W_{j}D_{j} \times C_{j}$ and comes either from the previous layer of the decoder (next section) or, for the lowest feature level, is learnt and is the same for all inputs.

\begin{figure*}
    \centering
    \includegraphics[width=0.9\textwidth]{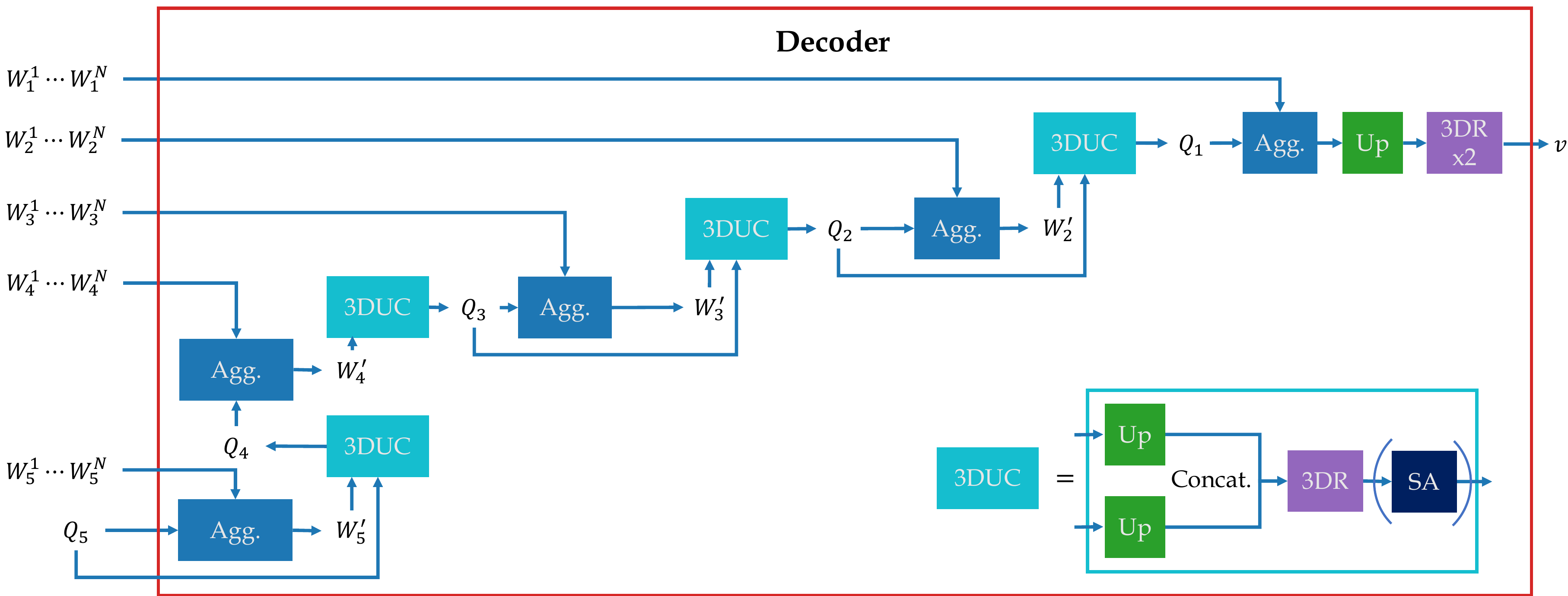}
    \caption{\textbf{Decoder} takes sets of feature maps at different levels and a learnt start query volume $Q_{5}$. Feature volumes from different images are aggregated, upscaled and passed through convolutional layers. After the features have been decoded, they are passed through a single convolutional layer to output the radiance field $\bv$. Self attention layers SA are applied when leading to $Q_{4}, Q_{3}, Q_{2}$, but not when outputting $Q_{1}$.}
    \label{fig:architecture_decoder}
\end{figure*}

\paragraph{Decoder and query volume.} 
The decoder takes $MN$ feature volumes as input: $N$ feature volumes $W_{j}^{1},\dots,W_{j}^{N}$ for each of $M$ feature levels $j$ (\cref{fig:architecture_decoder}, left). 
Additionally, it takes as input a learnt feature volume $Q_{M}$ which is identical for all inputs. 
At each feature level $j$, the decoder aggregates feature maps (\cref{fig:architecture_aggregator}) $W_{j}^{1},\dots,W_{j}^{N}$ using query volume $Q_{j}$ to output an aggregated feature volume $W_{j}^{'}$.
Feature volumes $Q_{j}$ and $W_{j}^{'}$ are upscaled, concatenated and passed it through a 3D ResNet Convolutional Block (\cref{fig:architecture_decoder}, bottom right).
Once feature maps from all levels have been aggregated, the feature volume of size $C_{1} \times H \times W \times D$ is passed through a small 5-layer convolutional upscaling subnetwork.
Finally, we apply a $1 \times 1 \times 1$ non-activated convolution layer which outputs the radiance field $\bv$ of shape $4 \times H \times W \times D$.

{\small\bibliographystyle{ieee_fullname}\bibliography{more,vedaldi_specific,vedaldi_general}}